%% file: main.tex
\documentclass[10pt, journal]{IEEEtran}

\usepackage{amsmath,amssymb,amsfonts}
\usepackage{graphicx}
\usepackage{pifont}

\usepackage{amssymb}
\usepackage{amsthm}
\usepackage{xcolor}
\usepackage[figuresright]{rotating}

\usepackage{graphicx}
\graphicspath{{/}{fig/}}

\usepackage{comment}
\usepackage{array}
\usepackage{textcomp}
\usepackage{gensymb}
\usepackage{xcolor}
\usepackage{multirow}
\usepackage{booktabs}
\usepackage{enumitem}

\usepackage{mathtools}
\usepackage{breqn}
\usepackage{float}
\usepackage[noend]{algpseudocode}
\usepackage[vlined, ruled, shortend]{algorithm2e}

\usepackage{caption}
\usepackage{subcaption}

\usepackage{mathtools}
\usepackage{float}
\usepackage{hyperref}

\usepackage{pgfplots}
\pgfplotsset{compat=1.7}
\usepgfplotslibrary{groupplots}

\usepackage{multirow,tabularx}
\usepackage[ancient]{flushend}

\usepackage{makecell}

\newcommand{\RN}[1]{%
  \textup{\uppercase\expandafter{\romannumeral#1}}%
}

\usepackage{xcolor}
\colorlet{orange_maxim}{green!10!orange!90!}

\usepackage[capitalise]{cleveref}

\newlength\figureheight
\newlength\figurewidth
\usetikzlibrary{shapes.misc}

\tikzset{cross/.style={cross out, draw=black, minimum size=2*(#1-\pgflinewidth), inner sep=0pt, outer sep=0pt},
cross/.default={1pt}}

\renewcommand{\baselinestretch}{0.985}

\IEEEaftertitletext{\vspace{-1.5\baselineskip}}

\title{
    \huge 
    Multi-Modal Lidar Dataset for Benchmarking General-Purpose 
    Localization and Mapping 
    Algorithms
}

\author{
    \IEEEauthorblockN{
        \vspace{1em}
        Li Qingqing\IEEEauthorrefmark{1},
        Yu Xianjia\IEEEauthorrefmark{1},
        Jorge Pe\~na Queralta\IEEEauthorrefmark{1},
        Tomi Westerlund\IEEEauthorrefmark{1} \\[+0.5em]
    }
    \IEEEauthorblockA{
        \normalsize
        \IEEEauthorrefmark{1}\href{https://tiers.utu.fi}{Turku Intelligent Embedded and Robotic Systems (TIERS) Lab, University of Turku, Finland}.\\
        Emails: \textsuperscript{1}\{qingqli, xianjia.yu, jopequ, tovewe\}@utu.fi\\[+6pt] 
    }
}
  
\begin{document}

\maketitle
\thispagestyle{empty}
\pagestyle{empty}

\input{sec/00_Abstract.tex}
\IEEEpeerreviewmaketitle

\input{sec/01_Intro.tex}

\input{sec/02_RelatedWorks}

\input{sec/03_Methodology}

\input{sec/04_Experiments}
\input{sec/05_Conclusion}


\section*{Acknowledgment}

This research work is supported by the Academy of Finland's AeroPolis project (Grant 348480) and the Finnish Foundation for Technology Promotion (Grants 7817 and 8089).

\bibliographystyle{IEEEtran}
\bibliography{bibliography}

\end{document}

%% file: sec/00_Abstract.tex
\begin{abstract}
    %
    Lidar technology has evolved significantly over the last decade, with higher resolution, better accuracy, and lower cost devices available today. In addition, new scanning modalities and novel sensor technologies have emerged in recent years. 
    Public datasets have enabled benchmarking of algorithms and have set standards for the cutting edge technology. However, existing datasets are not representative of the technological landscape, with only a reduced number of lidars available. This inherently limits the development and comparison of general-purpose algorithms in the evolving landscape.
    %
    This paper presents a novel multi-modal lidar dataset with sensors showcasing different scanning modalities (spinning and solid-state), sensing technologies, and lidar cameras. The focus of the dataset is on low-drift odometry, with ground truth data available in both indoors and outdoors environment with sub-millimeter accuracy from a motion capture (MOCAP) system. For comparison in longer distances, we also include data recorded in larger spaces indoors and outdoors.
    The dataset contains point cloud data from spinning lidars and solid-state lidars. Also, it provides range images from high resolution spinning lidars, RGB and depth images from a lidar camera, and inertial data from built-in IMUs.
    This is, to the best of our knowledge, the lidar dataset with the most variety of sensors and environments where ground truth data is available.
    This dataset can be widely used in multiple research areas, such as 3D LiDAR simultaneous localization and mapping (SLAM),  performance comparison between multi-modal lidars, appearance recognition and loop closure detection. The datasets are available at: https://github.com/TIERS/tiers-lidars-dataset.

\end{abstract}

\begin{IEEEkeywords}
    Multi-robot systems, Autonomous driving, dataset, computer vision, solid state LiDAR, SLAM
 
\end{IEEEkeywords}

%% file: sec/01_Intro.tex
  
\section{Introduction}\label{sec:introduction}
  
 
 
 
 

    Lidar sensors have become an essential part of many autonomous systems, from state-of-the-art self-driving stacks~\cite{kato2018autoware} to aerial robots~\cite{liu2022large}. Key factors motivating the adoption of lidars in such systems include their long-range, accurate detection ability in 3D and robust performance in a variety of scenarios and environmental conditions. Lidar odometry, localization and mapping algorithms find applications in areas such as autonomous driving vehicles~\cite{li2020multi}, unmanned aerial vehicles~\cite{varney2020dales}, forest survey~\cite{yang2020individual}. 

\begin{figure}
    \centering 
    \begin{subfigure}{0.49\textwidth}
        \centering  
          \includegraphics[width=\textwidth]{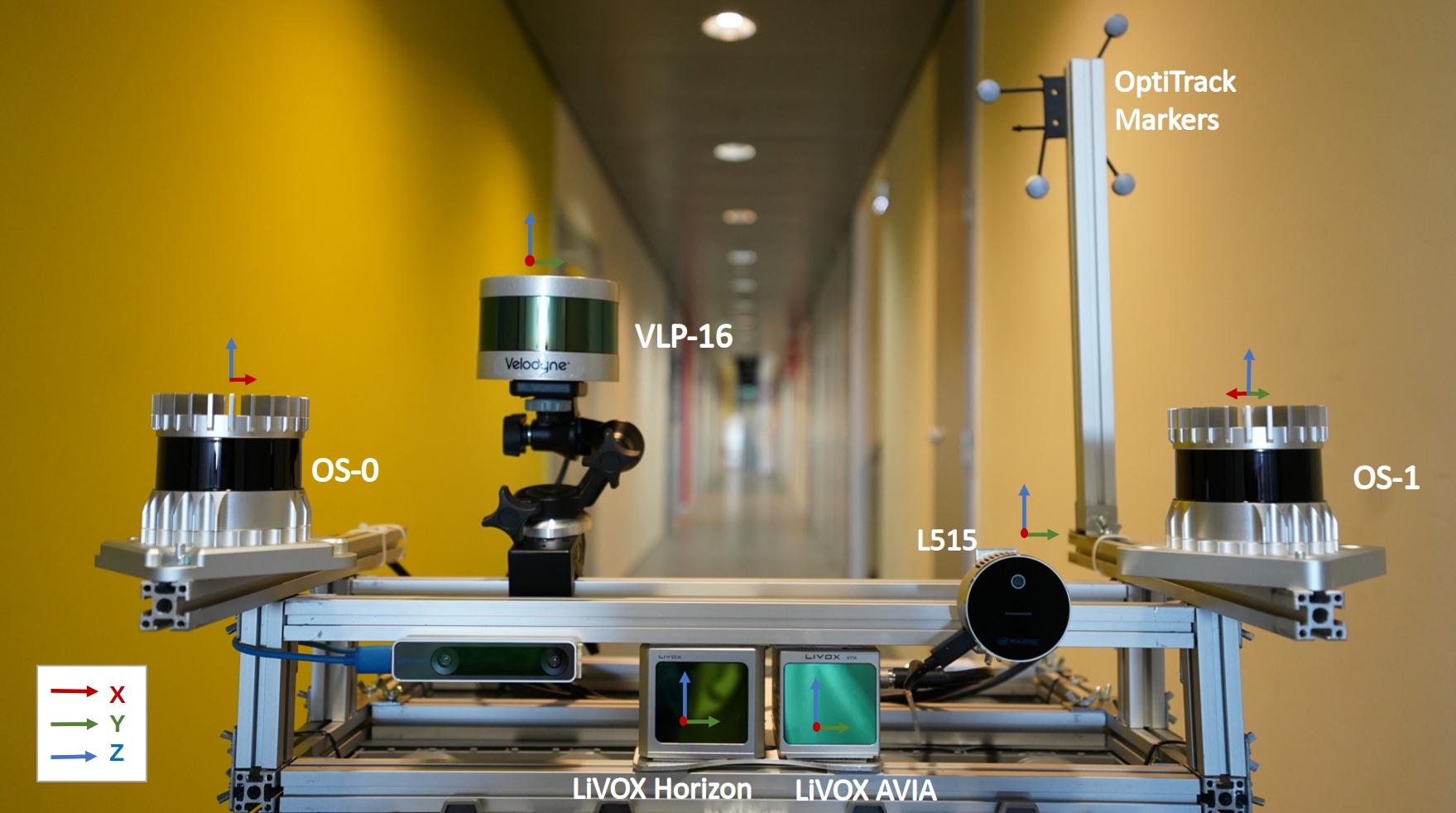}
        \caption{Front view of the multi-modal data acquisition system. Next to each sensor, we show the individual coordinate frames for the generated point clouds. 
        } 
        \label{fig:hardware_cfg}
    \end{subfigure}
    
    \vspace{1em}
    
    \begin{subfigure}{0.49\textwidth}
        \includegraphics[width=\textwidth]{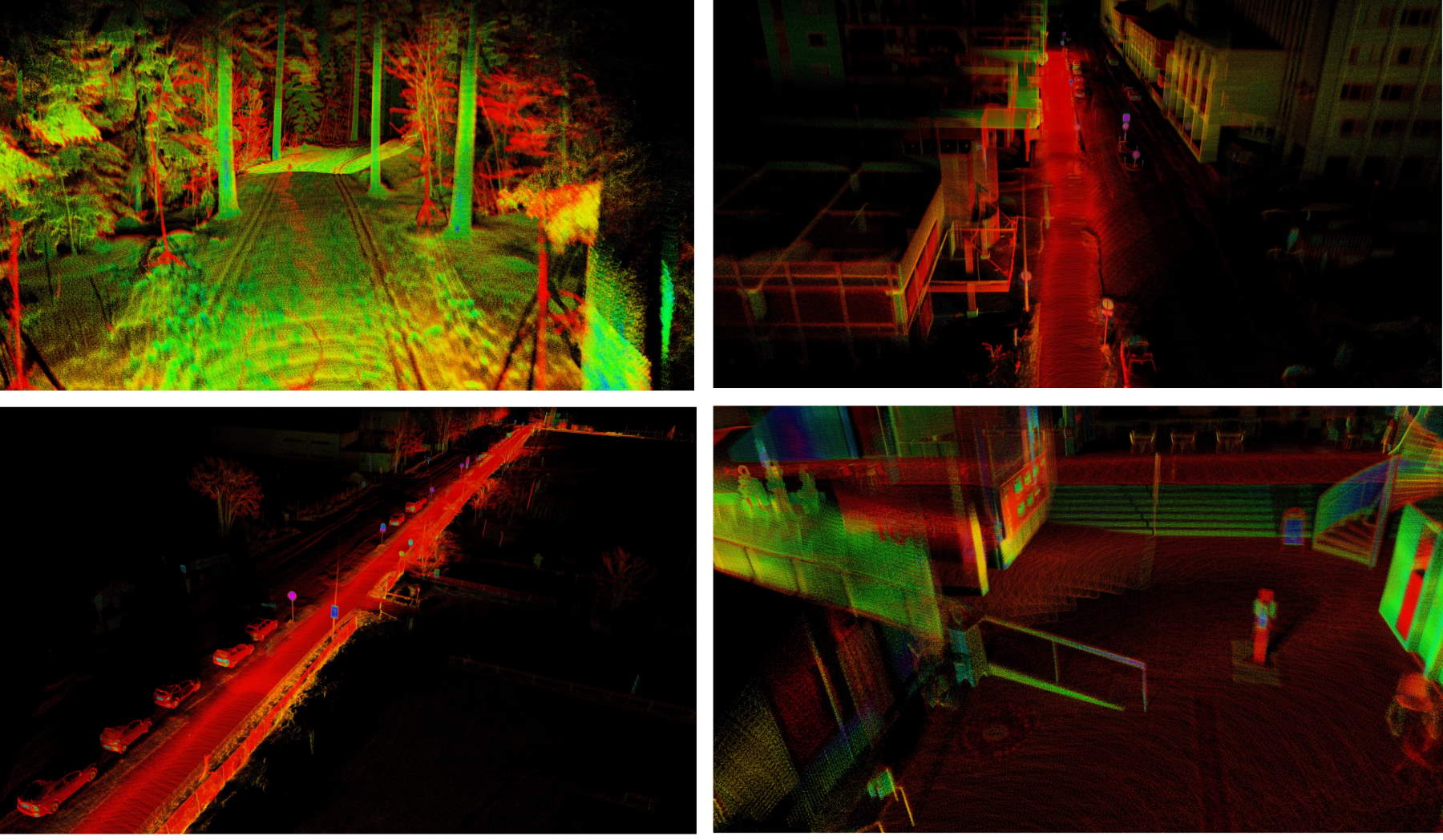}
        \caption{Samples of map data form different dataset sequences. From left to right and top to down, we display maps generated from a forest, an urban area, an open road and a large indoors hall, respectively.}
        \label{fig:dataset_env} 
    \end{subfigure}
    \caption{Multi-modal lidar data acquisition platform and samples from maps obtained in the different environments included in the dataset.}
\end{figure}

    High-resolution multi-beam spinning lidars have enabled high degrees of situational awareness in mobile robotic solutions. However, the larger number of vertical channels in a mechanically spinning lidar has meant until recently a cost too high for wide adoption in mobile platforms. While costs have significantly reduced in recent years as the technology matured, solid-state lidars able of generating high-density point clouds have emerged as an alternative. With a closer resemblance to visual sensors with limited field of view (FoV) and dense scene scanning, solid-state lidars offer high performance and relatively lower cost~\cite{lin2020loamlivox, li2020towards}. These sensors also offer non-repetitive scanning patterns allowing for higher resolution outputs than spinning lidars~\cite{liu2021lowretina}. Despite these benefits, their unique features challenge traditional lidar-based odometry, localization and mapping methods owing to the limited FoV, the irregular scanning patterns, the non-repetitive sensing and the need for different techniques to overcome motion-induced distortions. If the existing algorithms can be adapted to better support new types of lidar sensors, solid-state lidar have the potential to significantly benefit perception pipelines including object detection and tracking and scene understanding in different scenarios. We see particular potential in tasks in unstructured environments where geometric features are often too few and sparse compared to urban structured environment. While there are a large number of lidar datasets recorded with spinning lidar in diverse environments, there is a lack of multi-modal datasets enabling further adoption and understanding of different types of lidars and data processing algorithms.
 

 

    One of the main limitations of existing datasets that we aim to overcome is the lack of availability of data from solid-state lidars. These lidars often present limited FoV, owing to the lack of mechanical rotation, and different scanning modalities. In addition, compared to spinning lidars, non-repetitive patterns have emerged in different products, as well as different patterns even when they are repetitive. The limited FoV together with the different scan patterns pose significant differences for many of the standard lidar odometry, localization and mapping algorithms. As a result, we believe that more data is necessary to advance research towards general-purpose and sensor-agnostic lidar data processing algorithms. To bridge this gap, we present this novel multi-lidar dataset with spinning lidars of three different resolutions, two solid-state lidars with different FoV and scan pattern, and a small lidar-camera.
 



The main contributions of this work and the presented dataset are the following:
\begin{enumerate}
    \item A dataset with data from 5 different lidar sensors and one lidar camera in a variety of environments. This is, to our knowledge, the most diverse dataset in terms of lidar sensors for these environments. The dataset includes spinning lidars with 16 (Velodyne VLP-16), 64 (Ouster OS1-64) and 128 (Ouster OS1-128) channels and different vertical FoVs. Two different solid-state lidars (Livox Horizon and Livox Avia) with different scanning patterns and FoVs are also included. A lidar camera (RealSense L515) provides RGB images and lidar-aided depth images. Low-resolution images with depth, near-infrared and laser reflectivity data from the Ouster sensors complete the dataset. These are illustrated in Fig.~\ref{fig:hardware_cfg}.
    \item The dataset includes sequences with MOCAP-based ground truth in both indoors and outdoors environments. This is, to the best of our knowledge, the first lidar dataset to provide such accurate ground truth in forest environments in addition to indoor areas, albeit the limited trajectory length (see for samples Fig.~\ref{fig:dataset_env}).
    \item In addition to the MOCAP-labeled data, the dataset includes other sequences in large indoor halls, roads, and forest paths. The wide variety of sensors enables comparison between lidar odometry and mapping algorithms to an extent that was not possible before, with both general-purpose and sensor-specific approaches.
    \item Based on the presented dataset, we provide a baseline comparison of the state-of-the-art in lidar odometry, localization and mapping. We compare the odometry lift as well as the quality of the maps obtained with different sensors and different algorithms.
\end{enumerate}%

Based on the above characteristics of the presented dataset, we believe that it provides a timely and complimentary addition to existing datasets which are mostly focused towards mobile robots indoors or autonomous cars outdoors. We also show how the performance of the state-of-the-art lidar SLAM algorithms varies significantly based on the environment and the type of lidar sensor. The high degree of multi-modality in the sensor suite opens the door to research a variety of new challenges for the research community. We believe this dataset can aid in the design and development of future algorithms that better adapt to both structured and unstructured environments, to limited FoV of the sensors, and to different scanning modalities.


The structure of the paper is as follows. Section II surveys existing publicly available related datasets. Section III provides an overview of the configuration of the proposed sensor system. Section IV explains the details and specificity of the proposed data set. Finally, Section V concludes the study and introduces suggestions for further work.

%% file: sec/02_RelatedWorks.tex
\begin{table*}[t]  
 \centering
    \caption{Sensor specification for the presented dataset. Angular resolution is configurable in the OS1-64 (varying the vertical FoV). Livox lidars have a non-repetitive scan pattern that delivers higher angular resolution with longer integration times. Range is based on manufacturer information, with values corresponding to 80\% Lambertian reflectivity and 100 klx sunlight, except for the L515 lidar camera.} 
    \begin{tabular}{@{}lcccccccc@{}}  
    \toprule
        & IMU & Type & Channels & FoV & Angular Resolution & Range & Freq. & Points   \\
    \midrule   
        \textbf{Velodyne VLP-16} & N/A & spinning & 16  & 360°×30°    & V:2.0°, H:0.4°    & 100\,m      & 10\,Hz  & 300,000 pts/s \\  [0.5ex] 
        
        \textbf{Ouster OS1-64}      &  ICM-20948 & spinning & 64 & 360°×45°   & V:0.7°, H:0.18°    & 120\,m       & 10\,Hz  & 1,310,720 pts/s \\ [0.5ex] 
 
        \textbf{Ouster OS0-128}      & ICM-20948 & spinning & 128  & 360°×90°    & V:0.7°, H:0.18°    & 50\,m       & 10\,Hz  & 2,621,440 pts/s \\ [0.5ex]  
 
       \textbf{Livox Horizon}   &  BOSCH BMI088 & solid-state & N/A & 81.7°×25.1°          & N/A                 & 260\,m      & 10\,Hz  & 240,000 pts/s \\ [0.5ex] 
        
        \textbf{Livox Avia}      &  BOSCH BMI088 & solid-state & N/A & 70.4°×77.2°         & N/A                 & 450\,m      & 10\,Hz  & 240,000 pts/s\\ [0.5ex] 
 
        \textbf{RealSense L515}  &  N/A & lidar camera & N/A & 70°×55°          & N/A       & 9\,m        & 30\,Hz  & -  \\
     \bottomrule
    \end{tabular}
     \label{table:sensor_details}
\end{table*}


\section{Related Works}\label{sec:related_works}
 
    There are a number of datasets available that are relevant to this work, mostly gathered within the autonomous driving community. In Table~\ref{table:dataset_compare}, we compare the most relevant related datasets from the literature with ours. In the rest of this section, we address the key differences between the proposed dataset and existing ones.%

\subsection{Existing Datasets} 
    Datasets showcasing 3D lidar data and enabling benchmarking of approaches have had a significant impact within the research efforts in robust lidar odometry, localization and mapping algorithms. They have been particularly impactful within self-driving cars in the automotive industry~\cite{kato2018autoware}.
    One of the pioneers and perhaps the most significant dataset to date is arguably the KITTI benchmark\cite{geiger2013kitti}.
    The KITTI dataset includes a 64-beam 3D laser scanner, four gray-scale and color cameras, and a GPS/IMU navigation system within a single data-gathering platform. The KITTI benchmark has become an essential tool to evaluate the performance of algorithms in multiple tasks such as odometry, SLAM, object detection, or tracking, among others, in both academia and industry. Several similar datasets have also been published with a system composed of multiple cameras and spinning lidars, providing images and the corresponding point clouds in urban environment. Some relevant examples include the Oxford Robocar dataset~\cite{maddern2017oxford}, nuScences~\cite{caesar2020nuscenes}, or the EU long-term dataset~\cite{yan2020eu}. Multiple spinning lidars are often employed in these data collecting platform, albeit mostly sharing the same sensing modality or technology.%

    In addition to the myriad of datasets captured in urban road environments and focused towards research in autonomous driving, the literature also showcases efforts in off-road environments. 
    For example, the NCLT dataset provides a large-scale indoor and outdoor dataset with multi-modal sensors, including spinning lidars, cameras and IMU attached on a wheeled robot~\cite{carlevaris2016nclt}. 
    In another work, a handheld device comprised of one spinning lidar and depth camera was utilized to collect data from urban outdoor and vegetated environments~\cite{ramezani2020colleage}.
    With a larger variety of environments. a multi-sensor SLAM benchmark has been presented in~\cite{yin2021m2dgr}, with data captured in both indoor and outdoor environments. In relation to these works, we provide a wider variety of sensor data as well as more accurate ground truth in a selection of sequences. %

    There is also a number of datasets available in unstructured environments. For instance, the Robot Unstructured Ground Driving (RUGD) dataset captured from a small, unmanned mobile robot traversing in unstructured environments has been introduced in~\cite{wigness2019rugd}. The RUGD dataset contains different terrain types focusing on visual perception tasks like semantic segmentation. Several similar datasets in unstructured environments have been presented for tasks such as scene depth prediction~\cite{niu2020low}, terrain roughness understanding~\cite{gresenz2021off}, off-road pedestrian detection~\cite{pezzementi2018comparing}. Compared to these datasets, we provide MOCAP-based ground truth in a forest environment, while also including a wider variety of sensors.%

    In general, the number of publicly available datasets with solid-state lidar data is scarce. Among them, the PandaSet collects driving scenarios in urban environments with data from a forward-facing solid-state lidar and a 64-channels spinning lidar~\cite{xiao2021pandaset}. Additionally, Lin~\textit{et al.} presented an outdoor and indoor dataset with a solid-state lidar in college environment to test a novel lidar odometry and mapping (LOAM) algorithm tailored to solid-state lidar sensors~\cite{lin2020loamlivox}. In the present dataset, we provide a significantly higher number of sensors as well as ground truth both indoors and outdoors.
    

\subsection{SLAM with solid-state-lidar} 
 
    One of the key limitations of lidar technology preventing more widespread adoption for localization and mapping in mobile robots is the high cost of the sensors, specially compared to vision sensors. However, with lower-cost models becoming available, mainly solid-state lidars, multiple research efforts have been directed towards optimizing existing algorithms for the new sensing modalities and scanning patterns. 

    A robust, real-time LOAM algorithm for solid-state-lidar with small FoV and irregular samplings has been presented in~\cite{lin2020loamlivox} to address several fundamental challenges arising from solid-state-lidars. In another work, Lin \textit{et al.} proposed a decentralized framework for SLAM tasks with multiple solid-state-lidars to  increase the FoV and improve overall system robustness~\cite{lin2020decentralized}.      
    Inspired by local Bundle Adjustment (BA) techniques utilized in visual SLAM, a BA approach with an adaptive voxelization method to search feature correspondence and solve the problem of sparse features points in three-dimensional lidar data was presented in~\cite{liu2021balm}.
    More recent works have also worked towards improving lidar-based SLAM system robustness, with tightly-coupled lidar-inertial odometry and mapping schemes for both solid-state and mechanical lidars presented in~\cite{li2020towards, xu2021fastlio}. Additionally, a camera and solid-state lidar fusion SLAM framework have also been proposed in~\cite{zhu2021camvox}.

    In summary, the current research in the adaptation and tuning of algorithms for new lidar sensors lacks the support of a dataset for benchmarking and comparing the different approaches. Moreover, the lack of a truly heterogeneous and multi-modal dataset with various types of lidar sensors is preventing further comparisons between the methods to advance towards general-purpose lidar-based SLAM algorithms. To bridge these gaps, we focus on providing a dataset that can serve as an initial benchmark for odometry, localization and mapping in diverse environments and with different types of lidar sensors. We also hope that this dataset will further motivate research in the fusion of lidar data from different types of sensors.

%
 
\begin{figure} 
    \centering 
      
     \centering
     \includegraphics[width=0.45\textwidth]{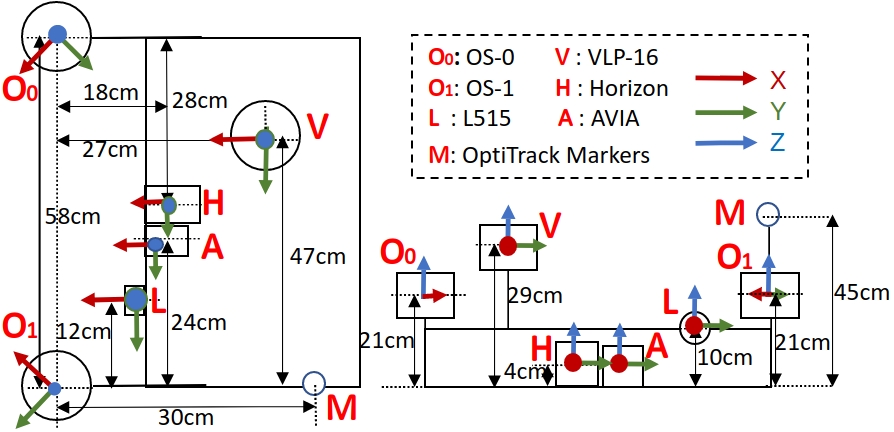}     
    \caption{Our data collecting platform, top view (left) and front view (right)} 
    \label{fig:device_scales}
\end{figure}

\begin{figure} 
    \centering 
      
     \centering
     \includegraphics[width=0.5\textwidth]{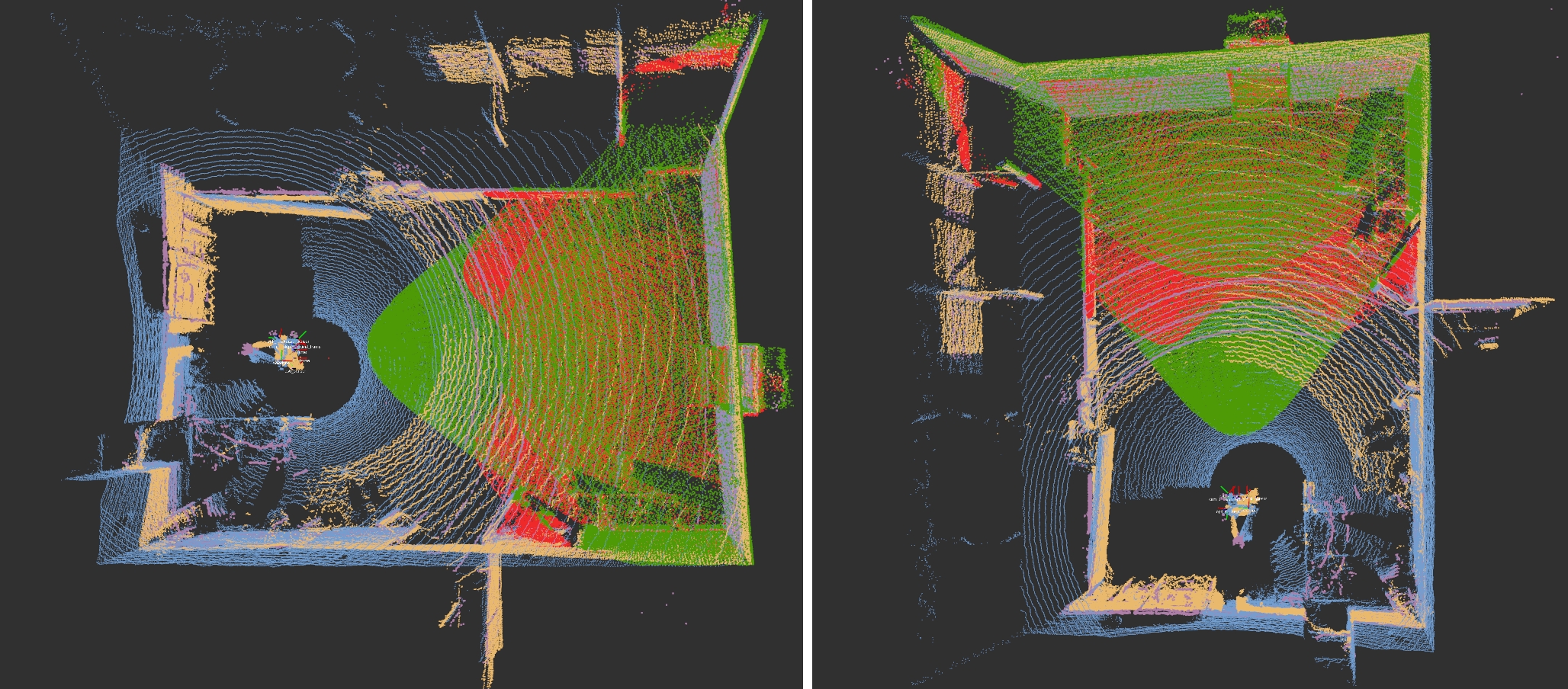}    
 
    \caption{Top view of point cloud data generated for the calibration process with multiple lidars. The red and green point clouds represent data obtained from the Livox Horizon and Avia, respectively. The purple, yellow, and blue clouds are from the VLP-16, OS1, and OS0 sensors.} 
    \label{fig:extrin_param}
\end{figure}

\begin{table}  
    \centering
    \caption{List of data sequences in our dataset (V: Velodyne VLP-16, H:Livox Horizon, A:Livox Avia, $O_0$: Ouster OS0, $O_1$: Ouster OS1)} 
    \begin{tabular}{@{}ccccc@{}}  
    \toprule 
        Sequence       & Description                  & Ground Truth       & Sensors             \\[0.5ex]
     \midrule   
        Forest01         & Forest(Winter,Square)       & Mocap          &V,H,A,$O_0$,$O_1$         \\ [0.5ex] 
        Forest02         & Forest(Winter,Straight)       & Mocap          &V,H,A,$O_0$,$O_1$         \\ [0.5ex]   
        Forest03         & Forest (long path)        & SLAM              &V,H                    \\[0.5ex]  
         
        Indoor01        & Office room(easy)         & Mocap     &V,H,A,$O_0$,$O_1$     \\[0.5ex] 
        Indoor02        & Office room(middle)       & Mocap     &V,H,A,$O_0$,$O_1$     \\  [0.5ex]  
        Indoor03        & Office room(hard)       & Mocap     &V,H,A,$O_0$,$O_1$     \\[0.5ex] 
        Indoor04        & Large Hall             & SLAM         &V,H,A,$O_0$,$O_1$     \\[0.5ex] 
        Indoor05        & Long Corridor        & SLAM         &V,H,A,$O_0$,$O_1$     \\    [0.5ex]   
        Road01          & Open road(short)            & SLAM       &V,H,A,$O_0$,$O_1$       \\  [0.5ex] 
        Road02          & Open road(long)            & SLAM       &V,H,A,$O_0$,$O_1$       \\  [0.5ex]  
    \bottomrule
    \end{tabular}
    \label{table:data_sequences}
\end{table} 

 
\begin{table*}[t]  
    \centering
    \caption{Comparison of related datasets with ours.} 
    
    \begin{tabular}{@{}cccccc@{}}  
     
    \toprule
        Dataset  & Year & Environment & Ground Truth &  LiDARs & Other   \\ [0.5ex]  
    \midrule
        KITTI\cite{geiger2013kitti}             & 2013   & \shortstack[c]{ Urban road}    & RTK\_GPS/INS & 3D-Velodyne HDL-64E @10\,Hz  & \shortstack{ 4× cameras , accel/gyro } \\ [0.5ex] 
     
        NCLT\cite{carlevaris2016nclt}  & 2017   & \shortstack[c]{ Urban Indoor\\Outdoor}    & GPS/INS     &  \shortstack[c]{ 3D-Velodyne HDL-64E@10\,Hz\\2× 2D-Hokuyo @10/40\,Hz}  &   camera\\ [0.5ex]

        Oxford RobotCar\cite{maddern2017oxford}  & 2017   & \shortstack[c]{Urban Road}    & GPS/INS     &  \shortstack[c]{  2× 2D-SICK @50\,Hz\\3D-SICK @12.5Hz}  & \shortstack{ 4 Camera;   accel/gyro} \\ [0.5ex] 
     
        RUGB Dataset\cite{wigness2019rugd}       & 2019  & \shortstack[c]{Unstructured \\outdoor}  & - & 3D-Velodyne HDL-32E @10\,Hz & \shortstack{ GPS\&IMU ; 3× cameras} \\ [0.5ex] 
     
        nuScences\cite{caesar2020nuscenes} & 2020  & \shortstack[c]{Urban Road}   & -  &  \shortstack[c]{    3D-32-Beams Lidar @20\,Hz } &  \shortstack{ 6x Camera (RGB);GPS\&IMU; \\ 5x Radar@13Hz } \\ [0.5ex]

        Newer Colleage\cite{ramezani2020colleage}& 2020  & \shortstack[c]{ Urban outdoor\\Vegetated} & 6DOF ICP   & 3D-Ouster-64 @10\,Hz &  \shortstack{ D435i (Infrared); accel/gyro} \\ [0.5ex] 
     
        PandaSet\cite{xiao2021pandaset} & 2021  & \shortstack[c]{Urban road} & - &  \shortstack{  3D-Hesai-Pandar64 @10\,Hz \\ 3D solid-state lidar@10\,Hz } & \shortstack{ 6x Cameras. GNSS\&IMU  } \\ [0.5ex] 
     
         M2DGR \cite{yin2021m2dgr} &  2022      & \shortstack[c]{Urban In/Outdoors}  & \shortstack{  Laser 3D tracker \\ RTK\_GPS/INS} &   3D VLP-32C @10\,Hz  &  \shortstack{ 3 Cameras. GNSS\&IMU  } \\ [0.5ex] 
         
        Our Dataset                              & 2022  & \shortstack[c]{ Urban indoor\\ Urban road \\Forest}    & \shortstack[c]{  6DOF MoCAP \\ SLAM} & \shortstack[c]{3x 3D-Spinning lidar(16,64,128) @10\,Hz \\ 2x 3D-Solid-State-lidar @10\,Hz \\  LiDAR-Camera @30\,Hz}&
            \shortstack[c]{ 2x accel/gyro @200\,Hz \\ 2x accel/gyro @100\,Hz} \\  [0.5ex] 
    
         \bottomrule 
    \end{tabular}
    \label{table:dataset_compare}
\end{table*}

%% file: sec/03_Methodology.tex
\begin{figure}[t]
     \centering
     \includegraphics[trim={0cm 6cm 2cm 0cm}, width=0.49\textwidth]{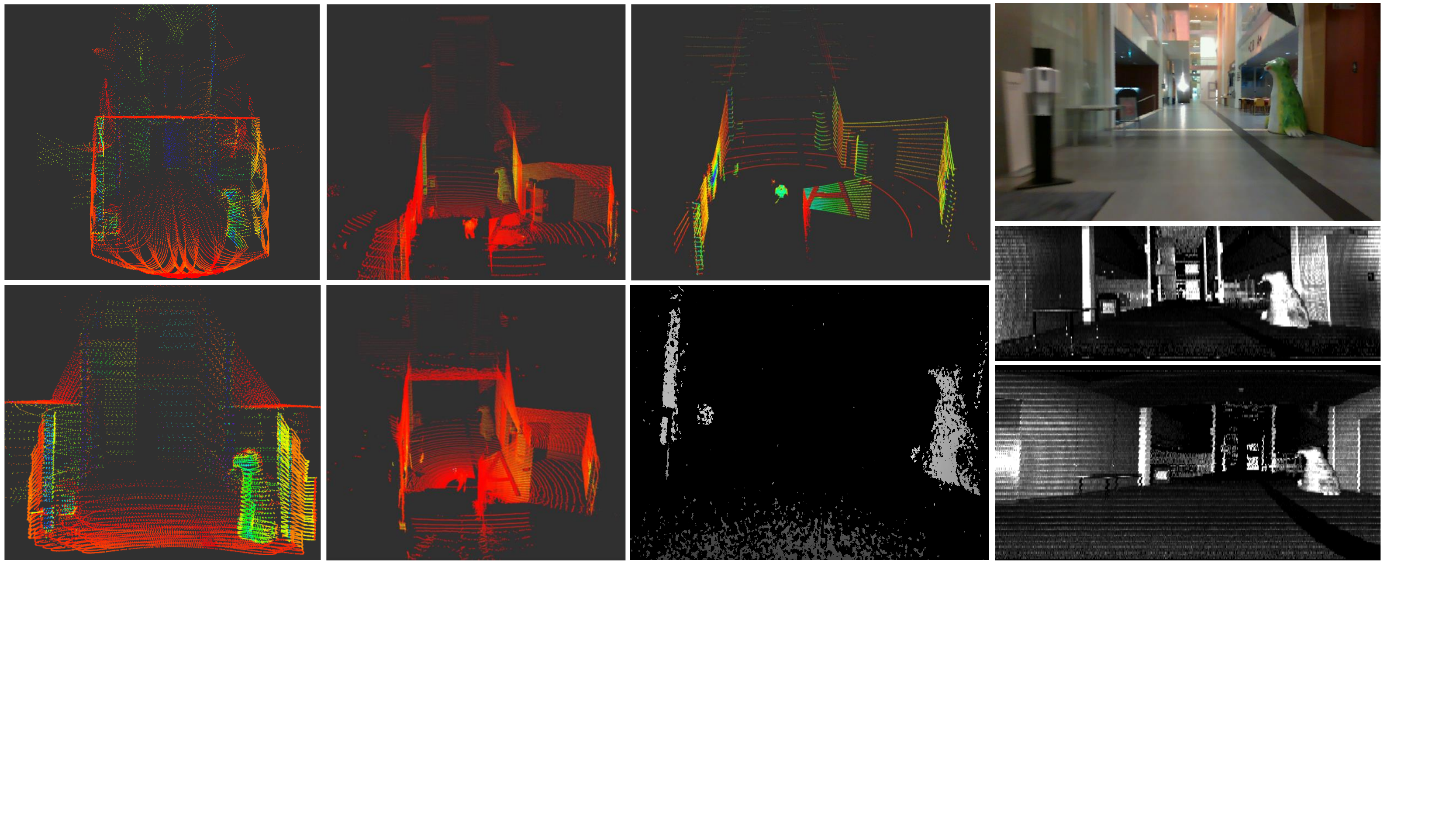}    
    \caption{Our dataset was captured by a rich suite of sensors. Subsets of the data from the \textit{Indoor04} sequence are visualized here. The leftmost column shows the lidar data from Avia and Horizon; the second column shows the lidar data from OS1 and OS0; the third column shows the data from the VLP-16 and depth image from L515. The rightmost column shows the RGB image from L515 and range images from 0S1 and OS0.} 
    \label{fig:data_samples}
\end{figure}
 
\section{System Overview}

    The sensor configuration of our data collection platform is shown in Fig.~\ref{fig:hardware_cfg}, and more specific information of each sensor is available in Table~\ref{table:sensor_details}. 
    Owing to the variety of environments where the platform has been used, it has been mounted on different types of mobile platforms. In road-like environments and large indoor halls, a Clearpath Husky mobile robot has been used. In forests outdoors with snow, it has been handheld. In small indoor spaces, it has been mounted on a mobile wheeled platform, manually pushed.
    In order to increase the usability of the dataset for benchmarking general-purpose algorithms, pitch and roll rotations have been applied in different configurations when handheld, in addition to standard horizontal settings where only the yaw angles varies if the surface where it is moving is horizontal.
    %
 
\subsection{Hardware}

    The core objective of the sensor system is to provide data from various lidar sensors with different characteristics, from novel low-cost solid-state lidar to 3D spinning lidars with different resolutions and vertical FoV, an including lidar cameras as well. 
    To this end, our data collecting platform includes three spinning lidars: 16-channels Velodyne lidar (VLP-16), 64-channels Ouster lidar (OS1), and 128-channels Ouster lidar (OS0). On the side of solid-state lidars, two units from Livox are installed: Horizon, with a FoV close to a rectangle, and Avia, with an almost-circular FoV. An Intel RealSense L515 lidar camera completes the setup.
    Regarding the physical configuration, the Horizon and Avia lidars were installed in the center of the frame facing forward. The L515 camera was attached to the front left of the platform. On the sides, the OS0 and OS1 sensors were mounted at a bit higher level, where the OS1 is turned 45 degrees clockwise, and the OS0 is turned 45 degrees anticlockwise. The Velodyne lidar is at the top-most position with the x-axis facing forward as well. Please refer to the top view of Fig.~\ref{fig:device_scales} for the detailed distances, positions and orientations. The Optitrack marker set for the MOCAP-based ground truth are fixed on the top of the aluminum stick to maximize its visibility and detection range.

    To ensure a low-latency and high-speed transmission of all data, the lidars are connected to a Gigabit Ethernet router and a computer onboard the platform featuring an Intel i7-10750h processor, 64\,GB of DDR4 RAM memory and 1\,TB SSD storage. The Optitrack system is also physically connected via Ethernet to the onboard computer on a separate interface to the lidars. Finally, the RealSense L515 camera is connected using a USB~3.0 port.
    
 \begin{figure}
     \centering   
     \includegraphics[trim={0cm 6cm 7cm 0cm}, width=0.45\textwidth]{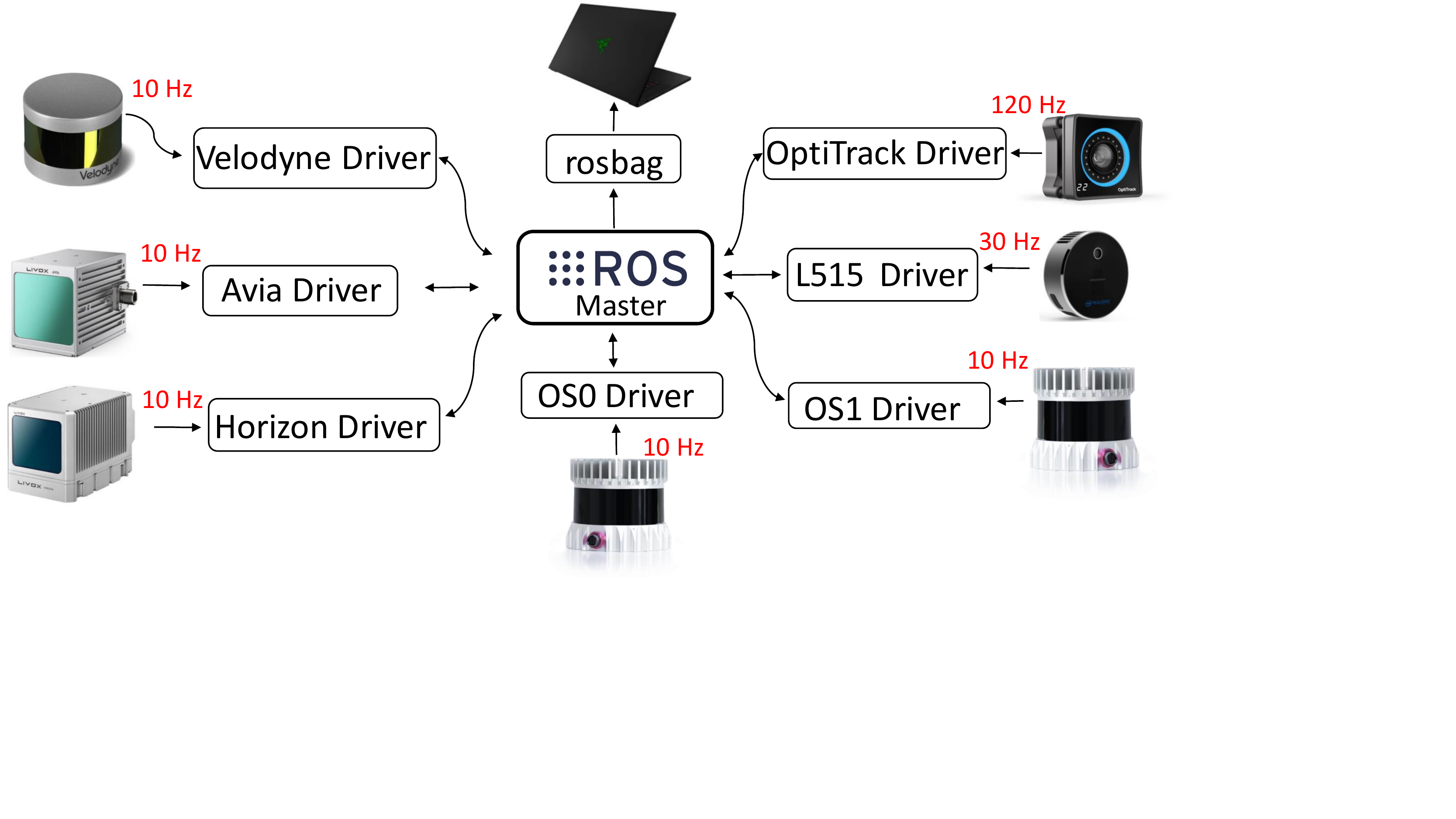}  
     \caption{ROS drivers and data gathering frequency for the different lidar sensors used in our platform.}
      \label{fig:software_cfg} 
\end{figure}

\subsection{Software} 
 
    Our software system is based entirely on ROS Melodic under Ubuntu 18.04. The set of ROS drivers and the publishing frequency of the different sensors is shown visually in Fig.~\ref{fig:software_cfg}. 
    Owing to the lack of hardware signals to synchronize the sensor data, as in other datasets in the literature~\cite{yan2020eu}, we approach the minimization of the data synchronization problem by running all the sensor drivers, data recording programs locally on a high performance computer. This, together with the networking equipment, aid in reducing the latency of data transmission at the hardware and software level (timestamped at the ROS drivers). In order to support a potentially wider use of the data, 
    the dataset also includes 
    the time stamp from built-in internal oscillators for both Livox and Ouster Lidars, and for both point cloud and IMU data, in addition to the timestamp included in the header of all ROS messages. We have also compared the angular velocities of IMUs together with data from the MOCAP system to conclude that the latency of our system is less than 10\,ms.
 

\subsection{Sensor Calibration} 
 
    The extrinsic parameters of the lidars are calculated based on optimization methods similar to those presented in~\cite{jeong2018complex}. We calculate the extrinsic parameters in an indoor office environment, while the sensor platform was stationary. The coordinate system of the Horizon lidar sensor is treated as the reference frame during the calibration process. Ten consecutive frames of point cloud data are integrated from the solid-state lidars to accumulate a higher degree of detail from the environment.
    The point cloud data from each different lidar is then transformed to the reference frame based on manual measurements of a set of features in the environment. Then, a Generalized Iterative Closest Point (GICP) method is employed to optimize the relative transformation between the reference frame and lidars iteratively~\cite{segal2009gicp}. For reference, in Fig.~\ref{fig:extrin_param} we show sample sensor data from one of the indoor environments after calibration.  
    The ROS package $livox\_camera\_lidar\_calibration$ was utilized to calibrate the extrinsic parameter between the Horizon sensor and the L515 lidar camera. The intrinsic parameters of lidars and the lidar camera are given based on factory settings and manufacturer information. A specific rosbag containing raw data recorded in stationary settings at the room shown in Fig.~\ref{fig:extrin_param} is provided for end-user re-calibration and potential application of different methods. 

\subsection{Groud truth}
 
    Generating accurate ground truth data in complex environments is a challenging task, as has been identified in multiple existing datasets. Many vehicular benchmarks utilize the pose generated from GNSS/INS fusion method as ground truth. However, multi-path effect can affect the accuracy of the pose estimated by GNSS sensors in forest and urban environments. For indoor environments, GNSS signals are unavailable.
    MOCAP systems have been widely adopted in indoor environments owing to their ability to provide millimeter-level accuracy in positioning data. However, the utilization of MOCAP systems is limited mainly by the range of the cameras, usually in the 10 to 20\,m range. The need for relatively complex setup of the system has also prevented the adoption of such systems for outdoors environments, and specially in unstructured environment such as forests.

    To meet the demands of reliable ground-truth data for diverse environments, the present dataset includes MOCAP-based ground-truth data in both a subset of indoor and forest environments. This enables millimeter level pose estimation as ground truth for odometry algorithms in both structured and unstructured environments, which can aid in researching low-drift odometry algorithms, accurate feature tracking, and reduction of motion-induced distortions in the data.
    For large-scale environment, where the MOCAP system is unavailable, we also provide location information as a reference from SLAM methods~\cite{xu2021fastlio}. In these settings, the higher-resolution lidar OS0-128 can be used as a baseline for the other sensors. We evaluate the SLAM algorithms in diverse environments, with a sample of environments shown in Fig.~\ref{fig:traj_samples}. From the different SLAM methods further characterized in the next section, those that use data from the OS0 sensor showcase the most robust performance in a series of sampled sequences.

\subsection{Data Sequences}
 
    The different subsets of out dataset are divided into three categories based on the environment: forest, indoor, urban outdoor. Table~\ref{table:data_sequences} lists all the sequences in our dataset. 
 
    Three sequences are provided for the forest environment. The forest data is collected at a forest in Turku, Finland $(60\degree28'14.3"N 22\degree19'54.8"E)$. The sequences Forest01 and Forest02 are collected in winter time with snow-covered ground. \textit{Forest01} includes a square-shaped trajectory, while in \textit{Forest02} the system is moved in a straight trajectory. Both of these sets include MOCAP data. A larger-scale forest recording is also provided in the \textit{Forest03} sequence, with Horizon and VLP-16 lidars mounted on a smaller, handheld device. These sequences can support research in areas from tree-counting to tree stem diameter estimation. The vast difference in environment structure from urban settings to forest settings can also support lidar-based general-purpose odometry, localization and mapping algorithms.

    The indoor environment then adds another dimension to the dataset with five data  sequences. The data is collected in rooms and open halls of ICT-City in Turku, Finland. Three sequences are collected in a large experiment room where data from the MOCAP system is available. From these, \textit{Indoor03} contains faster rotations and sudden movements, while positioning the sensors closer to objects in front and around. In consequence, most of the solid-state lidar view is blocked by objects or walls, presenting a significantly more challenging situation for odometry estimation algorithms based primarily on scan matching methods. The data in \textit{Indoor01} is recorded while maintaining a longer distance ($\approx50\,cm$) with objects and following a square-shaped trajectory with a reduced number of rotations. The \textit{Indoor02} sequence then features a circular trajectory with more rotation but again maintaining an even larger distance to objects than in \textit{Indoor03}. Sequences \textit{Indoor04} and \textit{Indoor05} correspond to recordings in  a large hall and long corridor environment, respectively.

    Finally, two sequences of open-road environment around the ICT-City building in Turku, Finland, are also included in this dataset. The length of \textit{Road01} is over 50\,m, while the traversed length of the trajectory in \textit{Road02} is about 500\,m. 


\subsection{Data Format} 
 
    The data is collected in ROS and saved with the rosbag format, which has become a standard in the robotics research community. Sampled data frames from a subset of the sensors is shown in~\ref{fig:data_samples}. The detailed data format for each type included in the dataset is listed as follow:

\begin{enumerate} 

    \item \textbf{Point cloud from spinning lidars} from the three spinning lidars, namely VLP-16, OS0-128 and OS1-64. The sensor message type from spinning lidars is recorded as $sensor\_msgs::PointCloud$. Each point in the point cloud holds four values $(x, y , z, I)$, where $x,y,z$ represent the local Cartesian coordinates, and $I$ is the laser reflectance of the point measured.  
    
    \item \textbf{Point cloud from solid-state lidar} from the  two solid-state lidars, namely Avia and Horizon. The message type of these solid-state lidars in the rosbags is Livox's custom data format named $livox\_ros\_driver/CustomMsg$. The customized message keeps the first point's timestamp of each frame as the base time and then provides an offset time relative to the base time for each point. This is needed as the non-repetitive pattern does not allowed for a posteriori estimation, unlike the spinning lidars, in which we can estimate the time difference between points based on the settings of the mechanical parts. With this information, the de-skew process can then be conducted on the data to compensate for the distortion in the point cloud data caused by the sensor's egomotion~\cite{lin2020loamlivox}.
    We have maintained this message type that contains time information for each point for algorithms that include in the processing flow the de-skew of point cloud data and other related research. However, standard ROS messages simplify the visualization of the point cloud with tools such as Rviz, and provide a format that many other lidar processing algorithms relying on standard ROS messages use~\cite{li2020towards}. Therefore, we provide format conversion tools to transform the Livox custom message data to the ROS standard message type $sensor\_msgs::PointCloud$. Each point is then converted to a new one that holds five values $(x, y, z, I, C)$, where $x, y, z$ is the local Cartesian coordinate set, $I$ is the intensity of the point, and where the integer part of $C$ represents the line number and the decimal part the point timestamp.
    
    \item \textbf{Images from lidar camera}. The RealSense L515 lidar camera is configured to publish RGB images with a size of 1920×1080, and depth images with a size of 1024×768. The message type is $sensor\_msgs::Image$ at frequency 10 Hz. The depth estimations are aided by the built-in lidar sensor.
 
    \item \textbf{Images from high-resolution spinning lidar}. The two high resolution lidar from Ouster, OS0-128 and OS1-64, can output fixed-resolution range images, near-infrared images captured by the laser sensor, and signal images. In these, each pixel represents the distance from the sensor origin to the point, the strength of the light captured, and the object's reflectivity, respectively. The images are published at frequency of 10\,Hz. The image data is spatially correlated, with 16 bits per pixel and a linear photo response. The message type in the rosbags is the standard $sensor\_msgs/Image$.
   
    \item \textbf{Inertial data from spinning and solid-state lidars}. There are in total four built-in 6-axis IMU sensors with 3‑axis gyroscope and a 3‑axis accelerometer, one in each of the Ouster and Livox lidars. They publish data at a frequency of 100\,Hz in the former and 200\,Hz in the latter. The data type of IMU data in the rosbags is again ROS' standard $sensor\_msgs::Imu$.
    
    \item \textbf{Ground truth data}. The ground truth data from the MOCAP system is included in rosbags as $geometry\_msgs::PoseStamped$ messages. They are obtained from the computer driving the set of OptiTrack cameras through a VRPN connection.
    
\end{enumerate}





%% file: sec/04_Experiments.tex
 \begin{figure*}
    \centering   
    \includegraphics[ trim={0cm 7cm 3cm 0cm}, width=\textwidth]{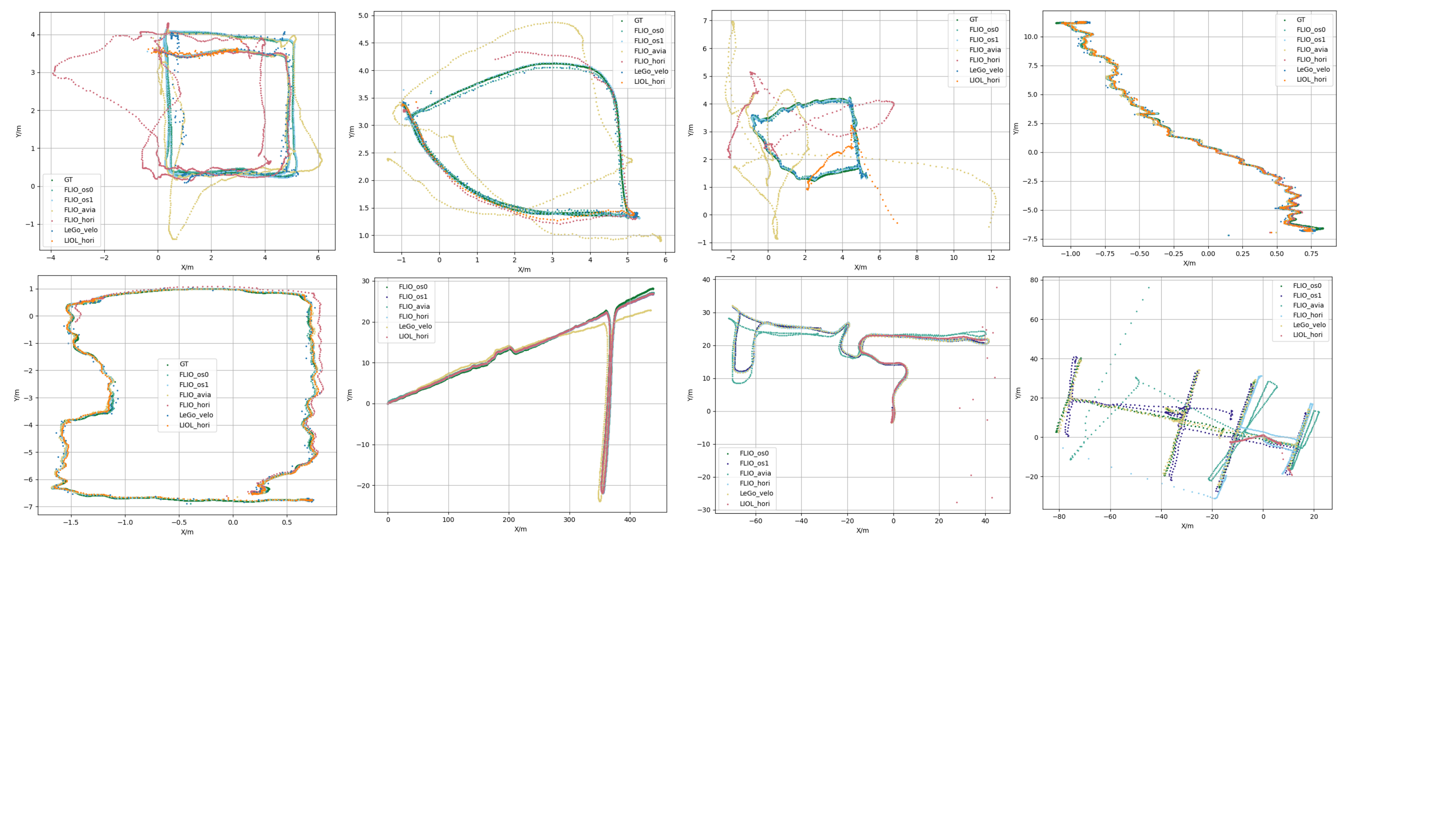}  
    \caption{Estimated trajectories. Top row: \textit{Indoor01}, \textit{Indoor02}, \textit{Indoor03}, \textit{Forest02}. Bottom row: \textit{Forest01}, \textit{Road02}, \textit{Indoor04}, \textit{Indoor05}.}
    \label{fig:traj_samples} 
\end{figure*}

\begin{table*}[t]
    \centering
    \caption{ ATE ($\mu/\sigma$) of selected SLAM methods (N/A when odometry estimations diverge). Best results for each sequence are in bold.} 
    \begin{tabular}{@{}ccccccc@{}}  
    \toprule 
        Sequence       & FLIO\_OS0       & FLIO\_OS1       & FLIO\_Avia     & FLIO\_Hori          & LeGo\_Velo       & LIOL\_Hori\\[0.5ex]
     \midrule     
        Indoor01        & \textbf{0.11 / 0.07}        &  0.12 / 0.04     &  0.58 / 0.3      & 0.65 / 0.24         & 0.22 / 0.19      & N/A                        \\[0.5ex] 
        Indoor02        & \textbf{ 0.17 / 0.12}     &  0.34 / 0.21        & 0.70 / 0.20       &  N/A          & 0.48 /  0.17   &  N/A            \\  [0.5ex]  
        Indoor03        & \textbf{0.16 / 0.09}      &  0.21 / 0.08        & N/A              & N/A            & 0.38 / 0.23     &  N/A                       \\[0.5ex]  
        Forest01        & 0.14 / 0.05      & 0.13 / 0.04    & 0.10 / 0.03    & 0.09 / 0.03          & 0.12 / 0.05     & \textbf{0.04/ 0.01}            \\  [0.5ex] 
        Forest02        & 0.13 / 0.07      & 0.12 / 0.06    & 0.09 / 0.03   & 0.11 / 0.05            & 0.31 / 0.05      &\textbf{0.07/ 0.04}          \\  [0.5ex]  
    \bottomrule 
    \end{tabular}    
    \label{table:ate_error}
\end{table*}

\begin{figure} 
    \centering   
    \begin{subfigure}{0.15\textwidth}
        \includegraphics[width=\textwidth, height=0.55\textwidth]{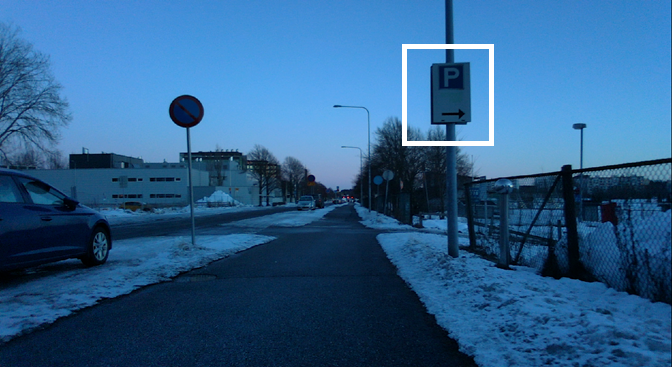}
        \caption{RGB}
    \end{subfigure}
    \begin{subfigure}{0.15\textwidth}
        \includegraphics[width=\textwidth, height=0.55\textwidth]{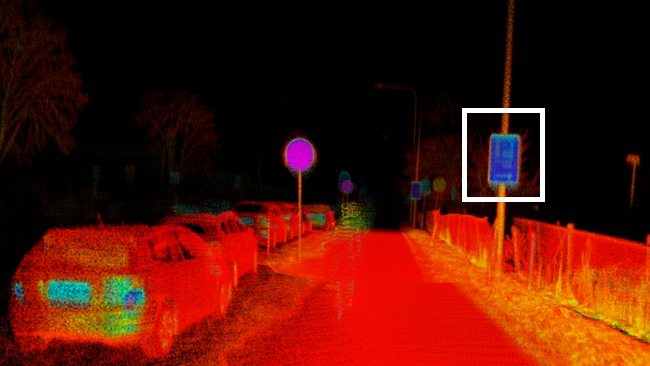}
        \caption{H-LIOL}
    \end{subfigure}
    \begin{subfigure}{0.15\textwidth}
        \includegraphics[width=\textwidth, height=0.55\textwidth]{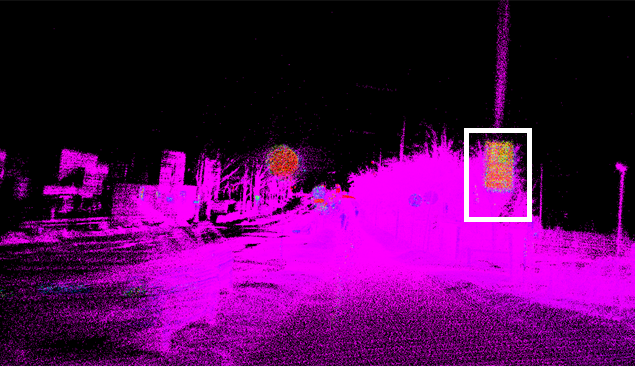}
        \caption{OS0-FLIO}
    \end{subfigure}
    \begin{subfigure}{0.06\textwidth}
        \includegraphics[width=\textwidth, height=1.3\textwidth]{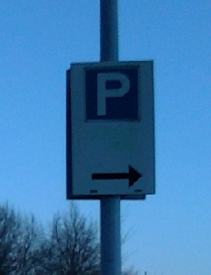}
        \caption{}
    \end{subfigure}
    \begin{subfigure}{0.06\textwidth}
        \includegraphics[width=\textwidth, height=1.3\textwidth]{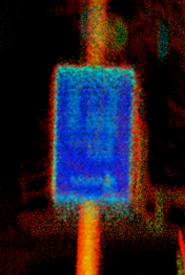}
        \caption{}
    \end{subfigure}
    \begin{subfigure}{0.06\textwidth}
        \includegraphics[width=\textwidth, height=1.3\textwidth]{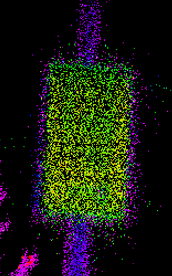}
        \caption{}
    \end{subfigure}
    \begin{subfigure}{0.06\textwidth}
        \includegraphics[width=\textwidth, height=1.3\textwidth]{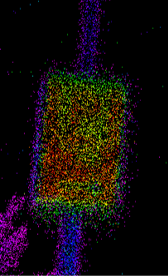}
        \caption{}
    \end{subfigure}
    \begin{subfigure}{0.06\textwidth}
        \includegraphics[width=\textwidth, height=1.3\textwidth]{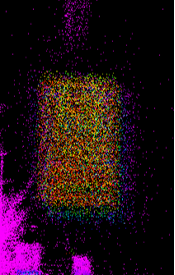}
        \caption{}
    \end{subfigure}
    \begin{subfigure}{0.06\textwidth}
        \includegraphics[width=\textwidth, height=1.3\textwidth]{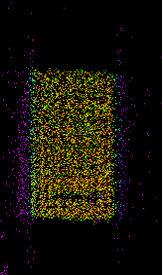}
        \caption{}
    \end{subfigure}
    \begin{subfigure}{0.06\textwidth}
        \includegraphics[width=\textwidth, height=1.3\textwidth]{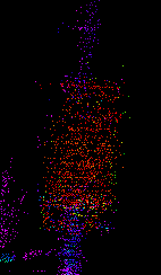}
        \caption{}
    \end{subfigure}
    \caption{Qualitative comparison of the mapping quality. Bottom row shows in (e) to (j) the     Horizon-based LIOL, Horizon, Avia, OS0, and OS1-based FLIO, and Velodyne's LeGo-LOAM maps, respectively.}
    \label{fig:map_details} 
\end{figure}

\section{Evaluation}\label{sec:methodology}
 
As a part of the dataset, we have evaluated several state-of-the-art SLAM algorithms on the different sequences. Through the rest of this section we discuss the best methods for different types of lidar data and environments.
 
\subsection{Lidar Odometry Benchmarking}
 
    Different lidar SLAM methods have been employed in our experiments. The FAST-LIO (FLIO) algorithm~\cite{xu2021fastlio}, a lidar-inertial odometry system that works for both spinning lidar and solid-state lidar, has been applied on Ouster lidars and Livox lidars leveraging the built-in IMUs. The objective here is to compare the performance of the same SLAM method applied to data from different types of lidars. In addition to FLIO, LeGo-LOAM\footnote{https://github.com/RobustFieldAutonomyLab/LeGO-LOAM} has been applied to data from the Velodyne lidar~\cite{shan2018lego}. Another SLAM system, LIO-LIVOX (LIOL)\footnote{https://github.com/Livox-SDK/LIO-Livox}, a tightly coupled SLAM that specifically developed for Horizon lidar, has also been tested on Horizon lidar data.

    The estimated trajectories are visualized in Fig.~\ref{fig:traj_samples}. 
    The plots are two-dimensional to improve readability as the changes in the vertical coordinate are minimal. Full data of the reconstructed paths is available in the dataset repository. From the results, one of the first conclusions is that the solid-state lidar-based SLAM system performs as well as or even better than the spinning lidars with the appropriate algorithms in the outdoors environments, but perform significantly more poorly in the indoor environments.

 
    For the outdoor sequences, \textit{Forest01}, \textit{Forest02}, and \textit{Road02}, all SLAM methods perform well, and the trajectories are completed without major disruptions.  For 
    For the indoor sequence \textit{Indoor01}, Avia- and Horizon-based FLIO are able to reconstruct the sensor trajectory but show that significant drift accumulates.
    In the same sequence, Horizon-based LIOL fails to reconstruct even the first loop in the trajectory. Similar behaviour is observed in the \textit{IndoorO2} sequence, with all the solid-state lidars failing completely in \textit{Indoor03}. In all of these sequences, all the methods applied to spinning lidars perform satisfactorily. This result can be expected as they have full view of the environment, which has a clear geometry.

    For the sequence \textit{Indoor04} showcasing a long corridor, all the spinning lidars can again reconstruct a complete trajectory. The best performance is obtained from OS0-based FLIO and Velodyne-based LeGo-LOAM, with correct alignment between the first and last location. However, angular drifts accumulates with OS1-based FLIO, while Horizon- and Avia-based algorithms result in diverging odometry estimations.

    In addition to the qualitative trajectory analysis, we also provide a quantitative analysis of the odometry error based on the MOCAP-based ground truth data
    %
    in Table~\ref{table:ate_error}. Absolute trajectory errors (ATE)~\cite{sturm2012benchmark} are employed as an evaluation metric. All trajectories are transformed to the local coordinate reference of the MOCAP markers, and aligned with global ground truth data reference. Then, we calculate ATE using the EVO toolset\,\footnote{https://github.com/MichaelGrupp/evo.git}. Methods based on spinning lidar data clearly show performance indoors, with naturally lower error as the vertical resolution increases. However, in the forest environment solid-state lidars demonstrate superior performance, with LIOL featuring an ATE error as low as 4\,cm mean error.

    In summary, the above results show that spinning lidars are more stable across different environments, while the solid-state lidars show significantly better cost-performance ratio in some outdoors environments.

\subsection{Mapping quality comparison}
 
    In the last part of our analysis, we compare the mapping quality generated from different lidars in urban open road environments as shown in Fig.~\ref{fig:map_details}. 
    From the figure, we can observe that the LIOL method applied to solid-state lidar presents the most detailed and clear map. It is worth noting that these maps have been generated with default configuration of the methods and without changing parameters such as the map update frequency. This result matches the quantitative results obtained with the same sensors and algorihtms in the forest environment. More results are available in our project page.

%% file: sec/05_Conclusion.tex
\section{Conclusion}\label{sec:conclusions}

    We have presented a novel dataset collected with a multi-modal lidar sensor system in diverse environments. The dataset includes data from lidars of different types (spinning and solid-state), resolution (16, 64 and 128 channels for spinning lidars) and scan patterns (for two different solid-state lidars), in addition to a lidar camera. This opens the door to future research in general-purpose algorithms, as our analysis shows that different algorithms clearly perform better in one or another type of lidar, if they are able to process the data at all. There is therefore a significant gap to be filled in more robust lidar odometry, localization and mapping algorithms. To aid in analyzing the drift of the algorithms, we have provided ground truth data both indoors and in a forest environment. For comparison of the mapping quality mainly, we also provide data from larger indoor halls and urban roads.
    Finally, we expect the dataset to support research in multi-sensor fusion and well as multi-modal lidar data fusion. The variety of data and environments set this dataset apart from the literature, setting the ground for benchmarking and quantitative comparisons of present and future lidar SLAM algorithms.
